\documentclass[10pt,twocolumn,letterpaper]{article}

\usepackage{cvpr}
\usepackage{times}
\usepackage{epsfig}
\usepackage{graphicx}
\usepackage{amsmath}
\usepackage{amssymb}
\usepackage{mathrsfs}
\usepackage{caption}
\usepackage{adjustbox}
\usepackage{multirow}
\usepackage{booktabs}

\usepackage{bm}

\usepackage[pagebackref=true,breaklinks=true,letterpaper=true,colorlinks,bookmarks=false]{hyperref}

\cvprfinalcopy 

\def\cvprPaperID{7923} 

\ifcvprfinal\fi
\begin{document}

\title{BachGAN: High-Resolution Image Synthesis from Salient Object Layout }

 \author{Yandong Li$^{1}$$^{*}$  \quad Yu Cheng$^{2}$ \quad Zhe Gan$^{2}$   \quad Licheng Yu$^{2}$ \quad Liqiang Wang$^{1}$   \quad Jingjing Liu$^{2}$\\
    $^{1}$University of Central Florida  \quad $^{2}$Microsoft Dynamics 365 AI Research \\
    {\tt\small \{liyandong,lwang\}@ucf.edu, \{yu.cheng,zhe.gan,licheng.yu,jingjl\}@microsoft.com }\\
    }


\twocolumn[{%
	\renewcommand\twocolumn[1][]{#1}%
	\maketitle
	\vspace{-10mm}
	\begin{center}
		\centering
		\includegraphics[width=0.96\linewidth,height=0.27\linewidth]{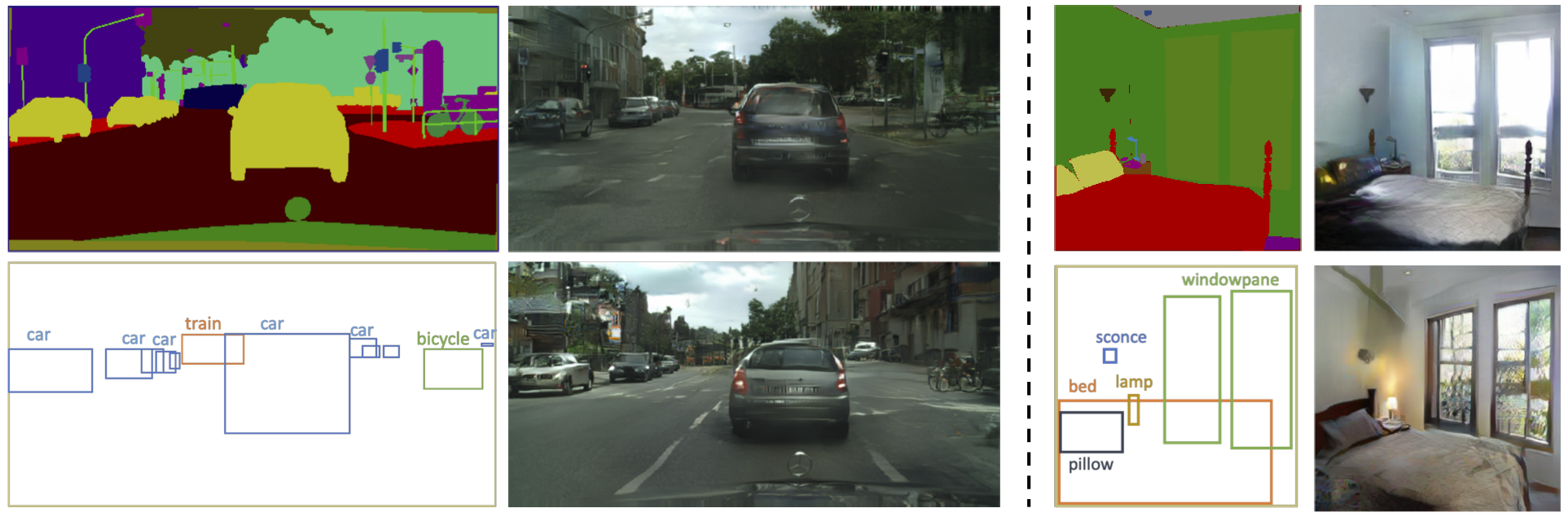}
		\vspace{-1mm}
		\captionof{figure}{Top row: images synthesized from semantic segmentation maps. Bottom row: high-resolution images synthesized from salient object layouts, which allows users to create an image by drawing only a few bounding boxes.}
		\label{fig:highlight}
	\end{center}
}]


\begin{abstract}
\vspace{-10pt}
We propose a new task towards more practical application for image generation - high-quality image synthesis from salient object layout. This new setting allows users to provide the layout of salient objects only (i.e., foreground bounding boxes and categories), and lets the model complete the drawing with an invented background and a matching foreground.
Two main challenges spring from this new task: ($i$) how to generate fine-grained details and realistic textures without segmentation map input; 
and ($ii$) how to create a background and weave it seamlessly into standalone objects. 
To tackle this, we propose Background Hallucination Generative Adversarial Network (BachGAN), which first selects a set of segmentation maps from a large candidate pool via a background retrieval module, then encodes these candidate layouts via a background fusion module to hallucinate a suitable background for the given objects. 
By generating the hallucinated background representation dynamically, our model can synthesize high-resolution images with both photo-realistic foreground and integral background. 
Experiments on Cityscapes and ADE20K datasets demonstrate the advantage of BachGAN over existing methods, measured on both visual fidelity of generated images and visual alignment between output images and input layouts.\footnote{Project page: https://github.com/Cold-Winter/BachGAN.}    
\end{abstract}
\vspace{-3pt}
{\let\thefootnote\relax\footnote{{$^{*}$ This work was done while the first author was an intern at Microsoft.}}}

\vspace{-4mm}
\section{Introduction}


Pablo Picasso once said ``\textit{Every child is an artist. The problem is how to remain an artist once grown up.}''
Now with the help of smart image editing assistant, our creative and imaginative nature can well flourish. 
Recent years have witnessed a wide variety of image generation works conditioned on diverse inputs, such as text~\cite{stackgan,attngan}, scene graph~\cite{sg2im}, semantic segmentation map~\cite{isola2017image, wang2018high}, and holistic layout~\cite{zhao2019image}.
Among them, text-to-image generation provides a flexible interface for users to describe visual concepts via natural language descriptions~\cite{stackgan,attngan}.
The limitation is that one single sentence may not be adequate for describing the details of every object in the intended image.

Scene graph \cite{sg2im}, with rich structural representation, can potentially reveal more visual relations of objects in an image.
However, pairwise object relation labels are difficult to obtain in real-life applications.
The lack of object size, location and background information also limits the quality of synthesized images.

Another line of research is image synthesis conditioned on semantic segmentation map.
While previous work~\cite{isola2017image,wang2018high,park2019semantic} has shown promising results, collecting annotations for semantic segmentation maps is time consuming and labor intensive.
To save annotation effort, Zhao et al.~\cite{zhao2019image} proposed to take as the input a holistic layout including both foreground objects (\emph{e.g.}, ``cat'', ``person'') and background (\emph{e.g.}, ``sky'', ``grass'').
In this paper, we push this direction to a further step and explore image synthesis given salient object layout only, with just coarse foreground\footnote{\emph{Salient} and \emph{foreground} are used interchangeably in this paper.} object bounding boxes and categories.
Figure~\ref{fig:highlight} provides a comparison between segmentation-map-based image synthesis (top row) and our setting (bottom row). 
Our task takes foreground objects as the only input, without any background layout or pixel-wise segmentation map.



The proposed new task presents new challenges for image synthesis: ($i$) how to generate fine-grained details and realistic textures with only a few foreground object bounding boxes and categories; and
($ii$) how to invent a realistic background and weave it into the standalone foreground objects seamlessly. Note that no knowledge about the background is provided; while in~\cite{zhao2019image}, a holistic layout is provided and only low-resolution ($64\times 64$) images are generated. In our task, the goal is to synthesize high-resolution ($512\times 256$) images given very limited information (salient layout only).


To tackle these challenges, we propose \emph{Background Hallucination Generative Adversarial Network} (BachGAN). 
Given a salient object layout, BachGAN generates an image via two steps: 
($i$) a background retrieval module selects from a large candidate pool a set of segmentation maps most relevant to the given object layout; ($ii$) these candidate layouts are then encoded via a background fusion module
to hallucinate a best-matching background. 
With this retrieval-and-hallucination approach, BachGAN can dynamically provide detailed and realistic background that aligns well with any given foreground layout.
In addition, by feeding both foreground objects and background representation into a conditional GAN (via a SPADE normalization layer~\cite{park2019semantic}), BachGAN can generate high-resolution images with visually consistent foreground and background. 
 
Our contributions are summarized as follows:
\begin{itemize}
    \vspace{-1.5mm}
    \item We propose a new task - image synthesis from salient object layout, which allows users to draw an image by providing just a few object bounding boxes. 
    \vspace{-2mm}
    \item We present BachGAN, the key components of which are a retrieval module and a fusion module, which can hallucinate a visually consistent background on-the-fly for any foreground object layout.
    \vspace{-2mm}
    \item Experiments on Cityscapes~\cite{Cordts2016Cityscapes} and ADE20K~\cite{zhou2017scene} datasets demonstrate our model’s ability to generate high-quality images, outperforming baselines measured on both visual quality and consistency metrics.
\end{itemize} 

\section{Related Work}
\begin{figure*}[t]
  \centering
  \includegraphics[width = 0.88\textwidth]{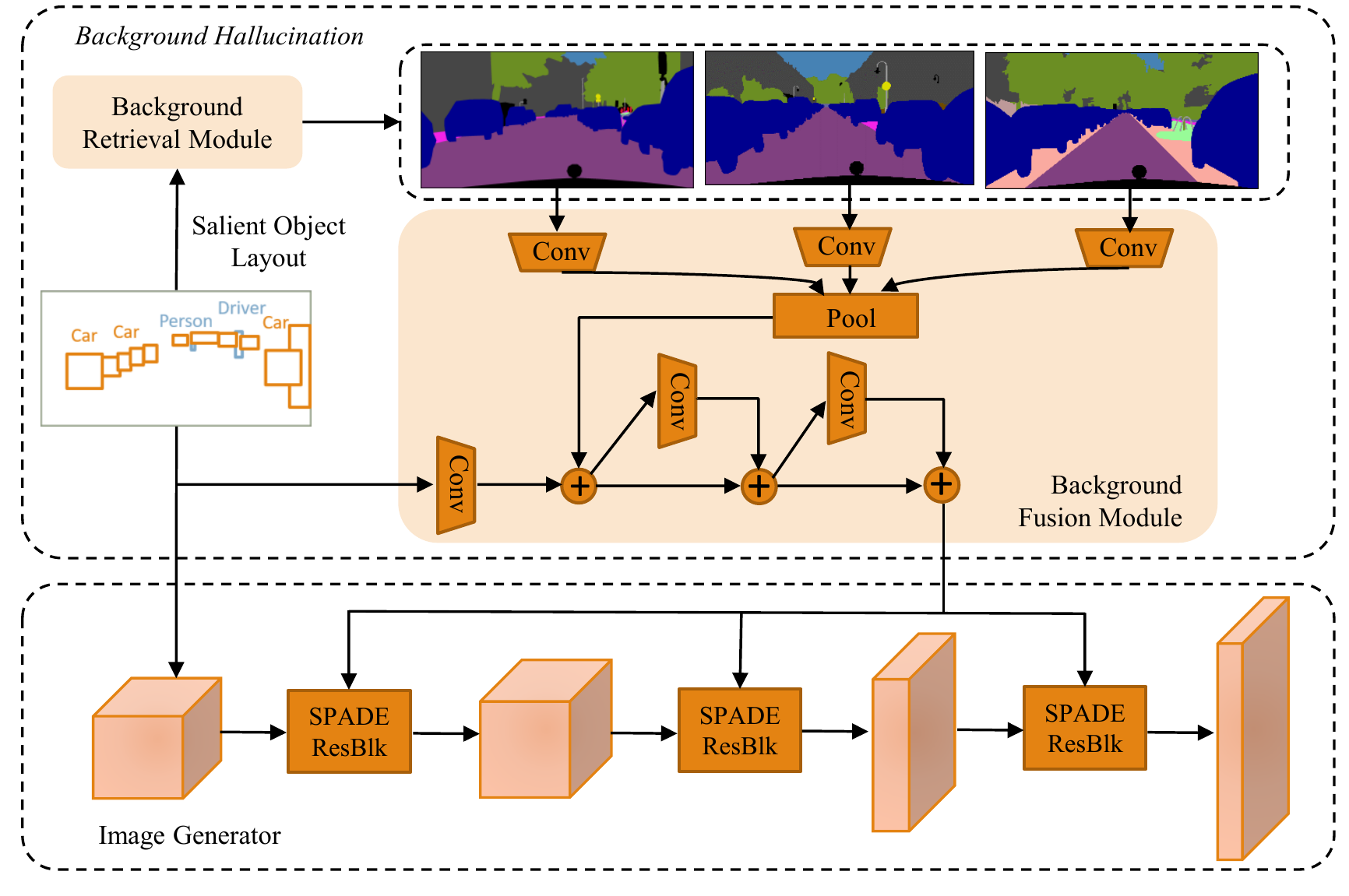}
  \vspace{-2mm}
  \caption{Overview of the proposed BachGAN for image synthesis from salient object layout.}
  \label{fig:overview}
  \vspace{-3mm}
\end{figure*}

\paragraph{Conditional Image Generation} Conditional image synthesis tasks can facilitate diverse inputs, such as source image \cite{isola2017image,Liu2017UIT,pathakCVPR16context,CycleGAN2017,NIPS2017_6650}, sketch \cite{sangkloy2016scribbler,NIPS2017_6650,xian2017texturegan}, scene graph \cite{sg2im,ashual2019specifying}, text \cite{mansimov16_text2image,stackgan,attngan,li2019object,li2019storygan}, video clip \cite{Lee2019CVPR,fan2019controllable,shen2019facial}, and dialogue \cite{SharmaSMKB18,seqattngan}. These approaches fall into three main categories: Generative Adversarial Networks (GANs) \cite{gan,mroueh2018sobolev}, Variational Autoencoders (VAEs) \cite{vae}, and autoregressive models \cite{pixelrnn,pixelcnn}. Our proposed model is a GAN framework aiming for image generation from salient layout only, which is a new task. 

In previous studies, 
the layout is typically treated as an intermediate representation between the input source (\emph{e.g.}, text~\cite{HongYCL18,li2019object} or scene graph~\cite{sg2im}) and the output image. 
Instead of learning a direct mapping from text/scene graph to an image, the model constructs a semantic layout (including bounding boxes and object shapes), based on which the target image is generated. Well-labeled instance segmentation maps are required to train the object shape generator. There is also prior work that aims to synthesize photo-realistic images directly from semantic segmentation maps~\cite{wang2018high,isola2017image}.
However, obtaining detailed segmentation maps for large-scale datasets is time consuming and labor intensive. 
In \cite{Lee2018CSP}, to avoid relying on instance segmentation mask as the key input, additional background layout and object layout are used as the input. ~\cite{zhao2019image} proposed the task of image synthesis from object layout; however, both foreground and background object layouts are required, and only low-resolution images are generated. 
Different from these studies, we propose to synthesize images from salient object layout only, which is more practical in real-life application where the user can simply draw the outlines of intended objects. 

\vspace{5pt}
\noindent\textbf{High-Resolution Image Synthesis} \,
Adversarial learning has been applied to image-to-image translation \cite{isola2017image,wang2018vid2vid}, to convert an input image from one domain to another using image pairs as training data.
$L_1$ loss \cite{Johnson2016Perceptual} and adversarial loss \cite{gan} are popular choices for many image-to-image translation tasks. 

Recently, Chen and Koltun \cite{ChenK17} suggest that it might be difficult for conditional GANs to generate high-resolution images due to training instability and optimization issues. To circumvent this, they use a direct regression objective based on a perceptual loss \cite{Dosovitskiy2016} and produce the first model that can synthesize high-quality images. Motivated by this, pix2pix-HD \cite{wang2018high} uses a robust adversarial learning objective together with a new multi-scale generator-discriminator architecture to improve high-resolution generation performance. In \cite{wang2018vid2vid}, high-resolution video-to-video synthesis are explored to model temporal dynamics. Park et al. \cite{park2019semantic} shows that spatially-adaptive normalization (SPADE), a conditional normalization layer that modulates the activations using input semantic layouts, can synthesize images significantly better than state-of-the-art methods. 

However, the input of all the aforementioned approaches is still semantic segmentation
map. In our work, we adopt the SPADE layer in our generator, but only use a salient object layout as the conditional input. This foreground layout is combined with hallucinated background to obtain a fused representation, which is then fed into the SPADE layer for image generation.

\section{BachGAN}
We first define the problem formulation and introduce preliminaries in Sec.~\ref{sec:task}, before presenting the proposed Background Hallucination Generative Adversarial Network (BachGAN). As illustrated in
Figure~\ref{fig:overview}, BachGAN consists of three components: ($i$) \emph{Background Retrieval Module} (Sec.~\ref{sec:retrieval}), which selects a set of segmentation maps from a large candidate pool given a foreground layout; ($ii$) \emph{Background Fusion Module} (Sec.~\ref{sec:fusion}), which fuses the salient object layout and the selected candidate into a feature map for background hallucination; and ($iii$) \emph{Image Generator}, which adopts a SPADE layer~\cite{park2019semantic} to generate an image based on the fused representation. Discriminators are omitted in Figure~\ref{fig:overview} for simplicity. 

\subsection{Problem Formulation and Preliminaries} \label{sec:task}
\paragraph{Problem Definition} Assume we have a set of images $\mathcal{I}$ and their corresponding salient object layouts $\mathcal{L}$. The goal is to train a model that learns a mapping from layouts to images, \emph{i.e.}, $\mathcal{L} \rightarrow \mathcal{I}$. Specifically, given a ground-truth image $\mathbf{I} \in \mathcal{I}$ and its corresponding layout $\mathbf{L} \in \mathcal{L}$, where $\mathbf{L}_i = (x_i, y_i , h_i , w_i)$ denotes the top-left coordinates plus the height and width of the $i$-th bounding box. Following~\cite{park2019semantic,wang2018high}, we first convert $\mathbf{L}$ into a label map $\mathbf{M} \in \{0,1\}^{H \times W \times C_o}$, where $C_o$ denotes the number of categories, and $H, W $ are the height and width of the label map, respectively. Different from the semantic segmentation map used in~\cite{wang2018high}, some pixels $\mathbf{M}(i,j)$ can be assigned to $n$ object instances, \emph{i.e.}, $\mathbf{M}(i, j) \in \{0, 1\}^{C_o} ~~ \textit{s.t.} ~~ \sum_{p} \mathbf{M}(i, j, p) = n$. 

\vspace{5pt}
\noindent\textbf{A Naive Solution} \, To draw in the motivation of our framework, we first consider a simple conditional GAN model and discuss its limitations. By considering the label map $\mathbf{M}$ as the input image, image-to-image translation models can be directly applied with the following objective:
\begin{align}\label{eq:gan}
\nonumber\min\limits_{G}\max\limits_{D} \,\, & \mathbb{E}_{\mathbf{M},\mathbf{I}}[\log (D(\mathbf{M},\mathbf{I}))] \\
+\,\, & \mathbb{E}_{\mathbf{M}}[\log (1-D(\mathbf{M},G(\mathbf{M})))]\,,
\end{align}
where $G$ and $D$ denote the generator and the discriminator, respectively. The generator $G(\cdot)$ takes a label map $\mathbf{M}$ as input to generate a fake image. 

State-of-the-art conditional GANs, such as pix2pix-HD~\cite{wang2018high}, can be directly applied here. However, some caveats can be readily noticed, as only a coarse foreground layout is provided in our setting, making the generation task much more challenging than when a semantic segmentation map is provided. Thus, we introduce Background Hallucination to address this issue in the following sub-section.

\subsection{Background Retrieval Module} \label{sec:retrieval}

The main challenge in this new task is how to generate a proper background to fit the foreground objects. Given an object layout $\mathbf{L}$ that contains $k$ instances: $\mathbf{L}_0^{C_0},\ldots,\mathbf{L}_k^{C_k}$, where $C_i$ is the category of instance $\mathbf{L}_i$, assume we have a memory bank $\mathbf{B}$ containing pairs of image $\mathbf{I}$ and its fine-grained semantic segmentation map $\mathbf{S}$ with $l$ instances: $\mathbf{S}_0^{C_0},...,\mathbf{S}_l^{C_l}$. We first retrieve a pair of $\mathbf{I}$ and $\mathbf{S}$ that contain the most similar layout to $\mathbf{L}$, by using a layout-similarity score, a variant of the Intersect over Union (IoU) metric, to measure the distance between a salient object layout and a fine-grained semantic segmentation map:
\begin{align}\label{eq:iouranking}
\text{IoU}_r &= \frac{\sum_{j=1}^{C}{\mathbf{S}^j } \bigcap \mathbf{L}^j }{\sum_{j=1}^{C}{\mathbf{S}^j } \bigcup \mathbf{L}^j } \,,
\end{align}
where $C$ is the total number of object categories, $\mathbf{S}^j = \bigcup \limits_{i} \mathbf{S}_i^j$, and $\mathbf{L}^j = \bigcup \limits_{i} \mathbf{L}_i^j$. $\bigcup$ and $\bigcap$ denote union and intersect, respectively. The proposed metric can preserve the overall location and category information of each object, since the standard IoU score is designed for measuring the quality of object detection. However, instead of calculating the mean IoU scores across all the classes, we use Eqn.~(\ref{eq:iouranking}) to prevent the 
weights of small objects from growing too high.

Given a salient object layout $\mathbf{L}_q$ as the query, we rank the pairs of image and semantic segmentation map in the memory bank by the aforementioned layout-similarity score.
As a result, we can obtain a retrieved image $\mathbf{I}_r$ with semantic segmentation map $\mathbf{S}_r$, which has a salient object layout most similar to the query $\mathbf{L}_q$. The assumption is that \textit{images with a similar foreground composition may share similar background as well}. Therefore, we treat the retrieved semantic segmentation map, $\mathbf{S}_r$, as the potential background for $\mathbf{L}_q$. Formally, we first convert the background of $\mathbf{S}_r$ into a label map $\mathbf{M}_{b}$: $\mathbf{M}_{b}(i, j) \in \{0, 1\}^{C_b} ~~ \textit{s.t.} ~~ \sum_{p} \mathbf{M}_{b}(i, j, p) = 1$, where $C_b$ denotes the number of categories in the background. Then, we produce a new label map, encoding both the foreground object layout and the fine-grained background segmentation map, by concatenating $\mathbf{M}_b$ and the foreground label map $\mathbf{M}_q$ of $\mathbf{L}_q$:
\begin{align}\label{eq:concat}
    \hat{\mathbf{M}} = [\mathbf{M}_b ; \mathbf{M}_q]\,,
\end{align}
where $[;]$ denotes concatenation, and the resulting label map is represented as $\hat{\mathbf{M}} \in \{0,1\}^{H \times W \times (C_o+C_b)}$. Note that the memory bank $\mathbf{B}$ can be much smaller than the scale of training data. Therefore, this approach does not require expensive pairwise segmentation map annotations to generate high-quality images.

\vspace{5pt}
\noindent\textbf{A Simple Baseline} \,
Based on the obtained new label map $\hat{\mathbf{M}}$, now we describe a simple baseline that motivates our proposed model. We consider the new label map as an input image, and adopt an image-to-image translation model. The following conditional GAN loss can be used for training:
\begin{align}
\label{eq:gan2d}
\nonumber \min\limits_{G}\max\limits_{D} \,\, &\mathbb{E}_{\hat{\mathbf{M}},\mathbf{I}_r}[\log (D(\hat{\mathbf{M}},\mathbf{I}_r))] 
 + \mathbb{E}_{\hat{\mathbf{M}},\mathbf{I}_q}[\log (D(\hat{\mathbf{M}},\mathbf{I}_q))] \\
+ \,\, & \mathbb{E}_{\hat{\mathbf{M}}}[\log(1-D(\hat{\mathbf{M}},G(\hat{\mathbf{M}})))]\,,
\end{align}
where $\mathbf{I}_q$ is the ground-truth image corresponding to the query $\mathbf{L}_q$, and $\mathbf{I}_r$ is the retrieved image.\footnote{Empirically, we observe that adding the retrieved image to the GAN loss improves the performance.} We name this baseline method ``\emph{BachGAN-r}''.
Though the ground-truth background cannot be obtained, the use of the retrieved background injects useful information that helps the generator synthesize better images, compared to the objective in Eqn. (\ref{eq:gan}). 

\begin{figure*}[t!]
  \centering
  \includegraphics[width = \textwidth]{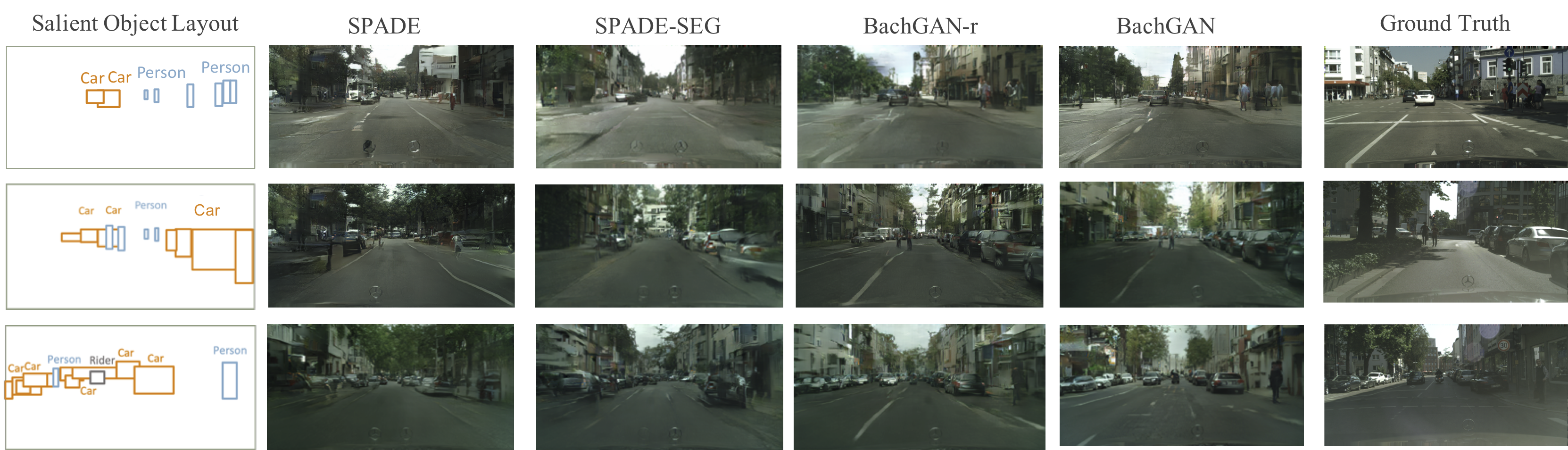}
  \caption{Examples of image synthesis results from different models on the Cityscapes dataset. Results from Layout2im are not included due to low resolution of the generated images (provided in Appendix). }
  \label{fig:vis:city}
  \vspace{-3mm}
\end{figure*}

\subsection{Background Fusion Module}\label{sec:fusion}
Though the retrieval-based baseline approach can hallucinate a background given a foreground object layout, the relevance of one retrieved semantic segmentation map to the input foreground layout is not guaranteed. One possible solution is to use multiple retrieved segmentation maps in Eqn.~(\ref{eq:gan2d}). However, this renders training unstable as the discriminator becomes unbalanced when several retrieved images are included in the loss function. More importantly, the dimension of the input label map is too high. In order to utilize multiple retrieved segmentation maps for a fuzzy hallucination of the background, we further introduce a Background Fusion Module to encode Top-$m$ retrieved segmentation maps to hallucinate a smoother background. 

Assume we obtain $m$ retrieved segmentation maps $\mathbf{S}_{r,0},...,\mathbf{S}_{r,m}$ with their corresponding background label maps $\mathbf{M}_{b,0},...,\mathbf{M}_{b,m}$. The query salient object layout $\mathbf{L}_q$ has a corresponding label map $\mathbf{M}_q$, where $\mathbf{M}_{b,i} \in \{0, 1\}^{H\times W \times C_b}$ and $\mathbf{M}_q \in \{0, 1\}^{H\times W \times C_o}$. As illustrated in Figure~\ref{fig:overview}, we first obtain $\hat{\mathbf{M}}_{r,0},...,\hat{\mathbf{M}}_{r,m}$ with Eqn.~(\ref{eq:concat}), where $\hat{\mathbf{M}}_{r,i} \in \{0,1\}^{H \times W \times (C_o+C_b)}$. $\mathbf{M}_q$ is padded with $\mathbf{0}$ to obtain a query label map $\hat{\mathbf{M}}_q$ with the same shape as $\hat{\mathbf{M}}_{r,i}$. $\hat{\mathbf{M}}_{r,0},...,\hat{\mathbf{M}}_{r,m}$ are then concatenated into $\hat{\mathbf{M}}_r \in \{0,1\}^{m \times H \times W \times (C_o+C_b)}$. A convolutional network $\mathcal{F}$ is then used to encode the label maps into feature maps:
\begin{align}
    \mathbf{m}_0 = \mathcal{F}(\hat{\mathbf{M}}_q) \oplus \text{Pool}(\mathcal{F}(\hat{\mathbf{M}}_r))\,,
\end{align}
where Pool represents average pooling, $\oplus$ denotes element-wise addition, and $\mathbf{m}_0 \in \mathbb{R}^{H \times W \times h}$ ($h$ is the number of feature maps). We then use another convolutional network $\mathcal{M}$ to obtain updated feature maps:
\begin{align}
    \mathbf{m}_t = \mathbf{m}_{t-1} \oplus \mathcal{M}(\mathbf{m}_{t-1})\,.
\end{align}
After $T$ steps, we obtain the final feature map $\hat{\mathbf{m}}=\mathbf{m}_T$, which contains information from both salient object layout and hallucinated background.

\vspace{5pt}
\noindent\textbf{Training Objective} \,
Based on the feature map $\hat{\mathbf{m}}$, BachGAN uses the following conditional GAN loss for training:
\begin{align}
\label{loss:gan2d}
\nonumber \min\limits_{G}\max\limits_{D} \,\, & \mathbb{E}_{\hat{\mathbf{m}},\mathbf{I}_q}[\log (D(\hat{\mathbf{m}},\mathbf{I}_q))] \\
+ \,\, & \mathbb{E}_{\hat{\mathbf{m}}}[\log(1-D(\hat{\mathbf{m}},G(\hat{\mathbf{m}})))]\,,
\end{align}
where $\mathbf{I}_q$ is the ground-truth image corresponding to the query $\mathbf{L}_q$. Compared with Eqn. (\ref{eq:gan2d}), multiple retrieved segmentation maps are used to hallucinate the background, which leads to better performance in practice.

\vspace{5pt}
\noindent\textbf{Image Generator} \,
Now, we describe how the generator $G(\cdot)$ takes $\hat{\mathbf{m}}$ as input to generate a high-quality image. In order to generate photo-realistic images, we utilize the spatially-adaptive normalization (SPADE) layer~\cite{park2019semantic} in our generator. Let $\mathbf{h}_i$ denote the activation feature map of the $i$-th layer of the generator $G$. Similar to batch normalization~\cite{ioffe2015batch}, SPADE~\cite{park2019semantic} first normalizes $\mathbf{h}_i$, then produces the modulation parameters $\bm\gamma$ and $\bm\beta$ to denormalize it, both of which are functions of $\hat{\mathbf{m}}$:
\begin{align}\label{eq: spade}
\hat{\mathbf{h}}_i = \text{norm}(\mathbf{h}_i) \otimes \bm\gamma(\hat{\mathbf{m}})   \oplus \bm\beta(\hat{\mathbf{m}})\,,
\end{align}
where $\hat{\mathbf{h}}_i$ denotes the output of a SPADE layer, norm $(\cdot)$ is a normalization operation, and $\otimes$ and $\oplus$ are element-wise production and addition, respectively. An illustration of the generator is provided in the bottom part of Figure~\ref{fig:overview}. More details about SPADE can be found in~\cite{park2019semantic}.

\begin{figure*}[t]
  \centering
  \includegraphics[width = 0.92\textwidth]{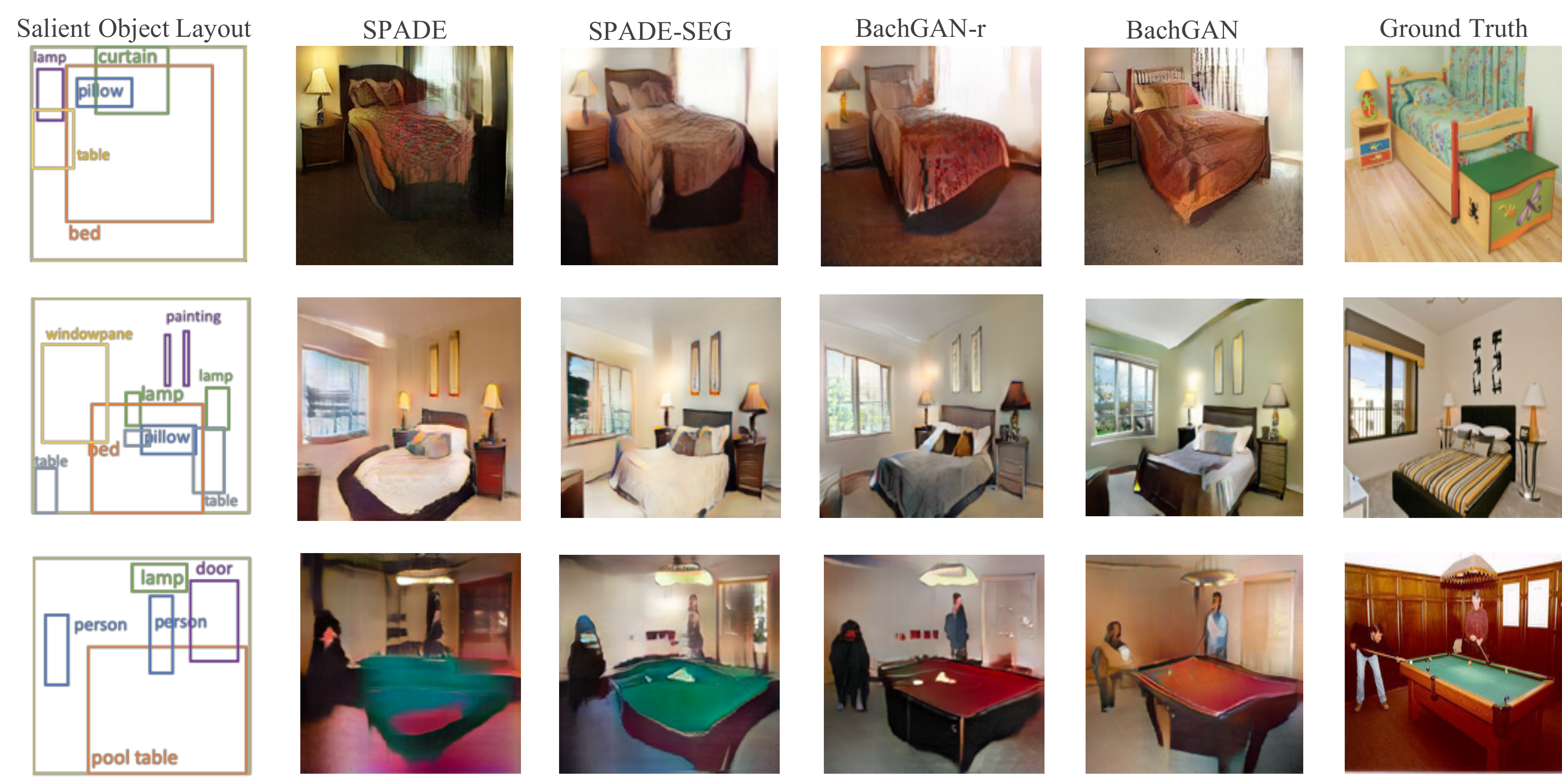}
  \caption{Examples of image synthesis results from different models on the ADE20K dataset. }
  \label{fig:vis:ade20}
\end{figure*}

\section{Experiments}
In this section, we describe experiments comparing BachGAN with state-of-the-art approaches on the new task, as well as detailed analysis that validates the effectiveness of our proposed model.

\subsection{Experimental Setup}

\noindent \textbf{Datasets} \, We conduct experiments on two public datasets: Cityscapes~\cite{Cordts2016Cityscapes} and ADE20K~\cite{zhou2017scene}. Cityscapes contains images with street scene in cities. The size of training and the validation set is 3,000 and 500, respectively. We exclude 23 background classes and use the remaining 10 foreground objects in the salient object layout. With provided instance-level annotations, we can readily transform a semantic segmentation instance to its bounding box, by taking the $\max$ and $\min$ of the coordinates of each pixel in an instance. ADE20K consists of 20,210 training and 2,000 validation images. The dataset contains challenging scenes with 150 semantic classes. We exclude the 35 background classes and utilize the remaining 115 foreground objects. There are no instance-level annotations for ADE20k, thus, we use a simple approach to find contours~\cite{suzuki1985topological} from a semantic segmentation map and then obtain the bounding box for each contour. A separate memory bank is used for each dataset. We train all the image synthesis methods on the same training set and report their results on the same validation set. 

\vspace{5pt}
\noindent \textbf{Baselines}  \,
We include several strong baselines that can generate images with object layout as input: 
\begin{itemize}
    \item \textbf{SPADE}: We adopt SPADE \cite{park2019semantic} as our first baseline, taking as input the salient object layout instead of semantic segmentation map used in the original paper. 
    \item \textbf{SPADE with Segmentation (SPADE-SEG)}: We obtain the second baseline by exploiting the pairs of segmentation mask and image from the memory bank. Besides GAN loss, the model is trained with an additional loss. It minimizes the segmentation loss between the real image and the output from the generator based on the memory bank.
    \item \textbf{Layout2im}: We use the code from  Layout2im~\cite{zhao2019image}, which generates images from holistic layouts and supports the generation of $64\times64$ images only.
\end{itemize}

\begin{table}[t!]
\begin{center}
\small
\begin{tabular}{c|cc|cc}
\hline
\multirow{2}{*}{Model} & \multicolumn{2}{c|}{Cityscapes} & \multicolumn{2}{c}{ADE20K} \\
\cline{2-5}
& Acc & FID & Acc & FID \\
\hline \hline
Layout2im~\cite{zhao2019image} & - & 99.1 & - & - \\

SPADE & 57.6 & 86.7 & 55.3 & 59.4 \\

SPADE-SEG & 60.2 & 81.2 & 60.9 & 57.2  \\

BachGAN-r & 67.3 & 74.4 & 64.5 & 53.2 \\

BachGAN & \textbf{70.4} & \textbf{73.3} & \textbf{66.8} & \textbf{49.8} \\
\hline
SPADE-v \cite{park2019semantic} & 81.9$^\dagger$ & 71.8$^\dagger$ & 79.9$^\dagger$ & 33.9$^\dagger$ \\
\hline
\end{tabular}
\end{center}
\vspace{-3mm}
\caption{Results on Cityscapes and ADE20K w.r.t. FID and the pixel accuracy (Acc). Results with ($\dagger$) are reported in \cite{park2019semantic}, serving as the upper bound of our model performance. }\label{tab:fid}
\vspace{-3mm}
\end{table}

\begin{figure*}[t]
  \centering
  \includegraphics[width = 0.95\textwidth]{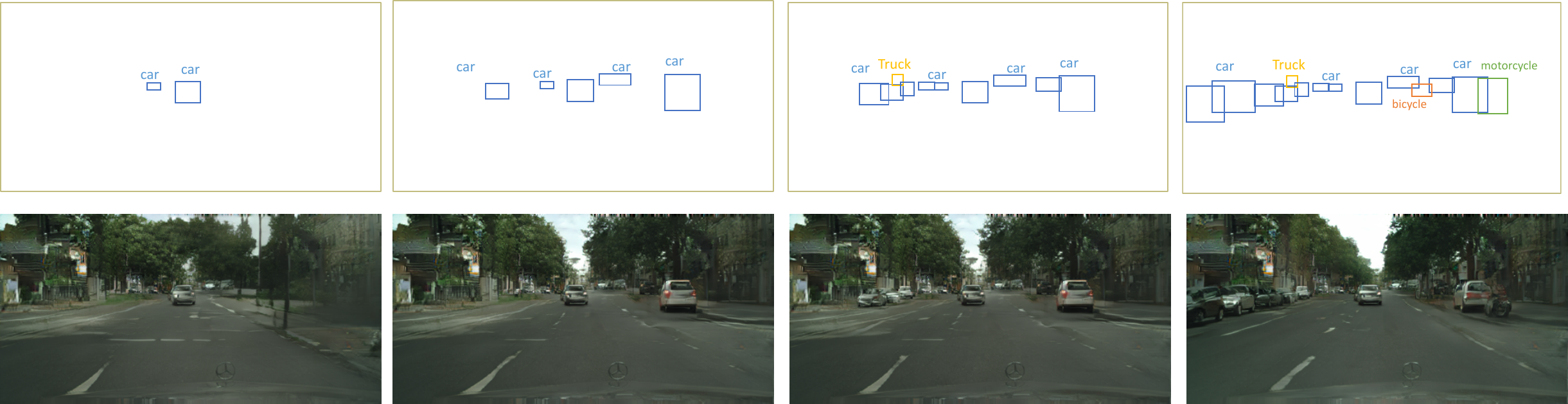}
  \caption{Generated images by adding bounding boxes sequentially to previous layout (Cityscapes).}
  \label{fig:seq_edit_cityscape}
  \vspace{-3mm}
\end{figure*}

\begin{figure*}[t]
  \centering
  \includegraphics[width = 0.88\textwidth]{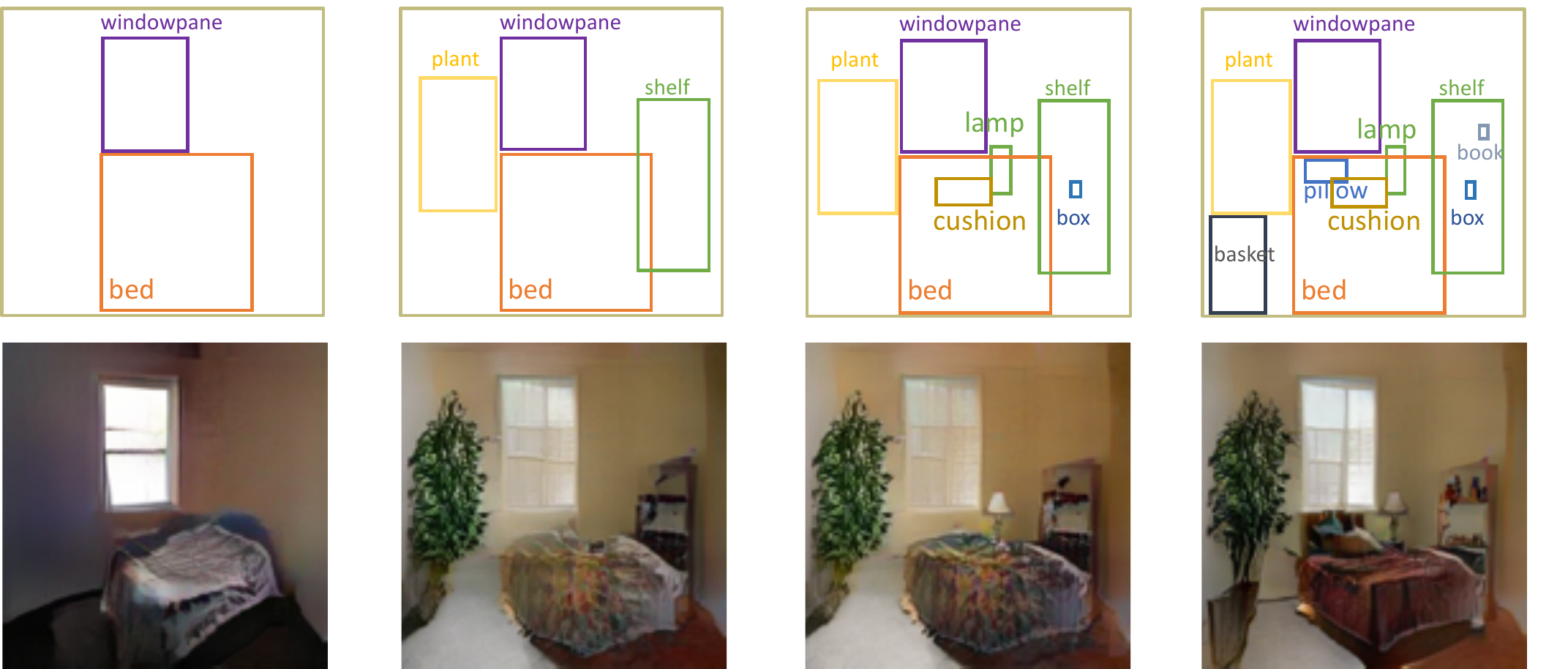}
  \caption{Generated images by adding bounding boxes sequentially to previous layout (ADE20k).}
  \label{fig:seq_edit_ade20k}
\end{figure*}

\begin{figure*}[t]
  \centering
  \includegraphics[width = 0.99\textwidth]{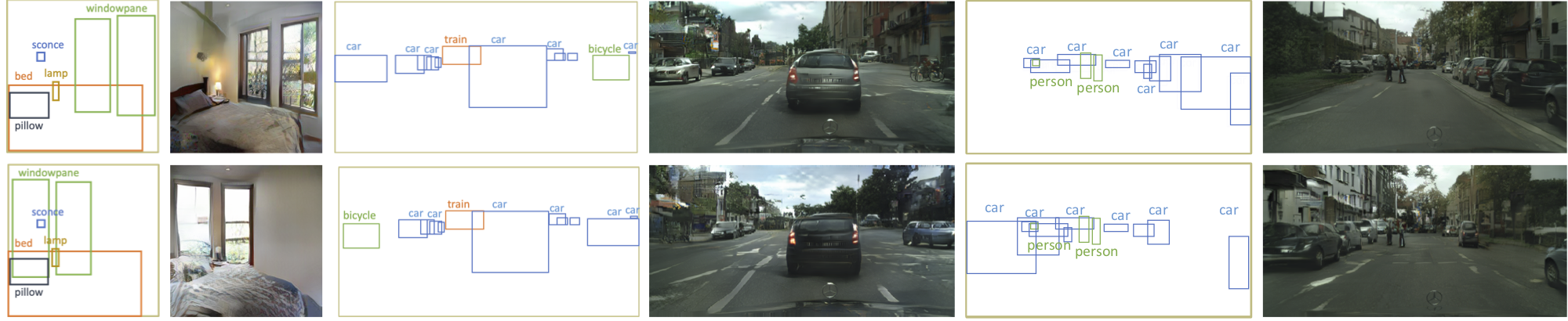}
  \caption{Top row: synthesized images based on salient object layouts from the test set. Bottom row: synthesized images based on salient layouts with flipping objects, modified from the top-row layouts. 
  }
  \label{fig:obj_flip}
  \vspace{-3mm}
\end{figure*}

\noindent \textbf{Performance metrics} \, Following \cite{ChenK17,wang2018high}, we run a semantic segmentation model on the synthesized images and measure the
segmentation accuracy. We use state-of-the-art segmentation networks: DRN-D-105 \cite{Yu2017} for Cityscapes, and UperNet101 \cite{xiao2018unified} for ADE20K. Pixel accuracy (Acc) is compared across different models. This is done using real objects cropped and resized from ground-truth images in the training set of each dataset. In addition to classification accuracy, we use the Frechet Inception Distance (FID) \cite{fid} to measure the distance between the distribution of synthesized results and the distribution of real images.

\begin{figure*}[t]
  \centering
  \includegraphics[width = 0.95\textwidth]{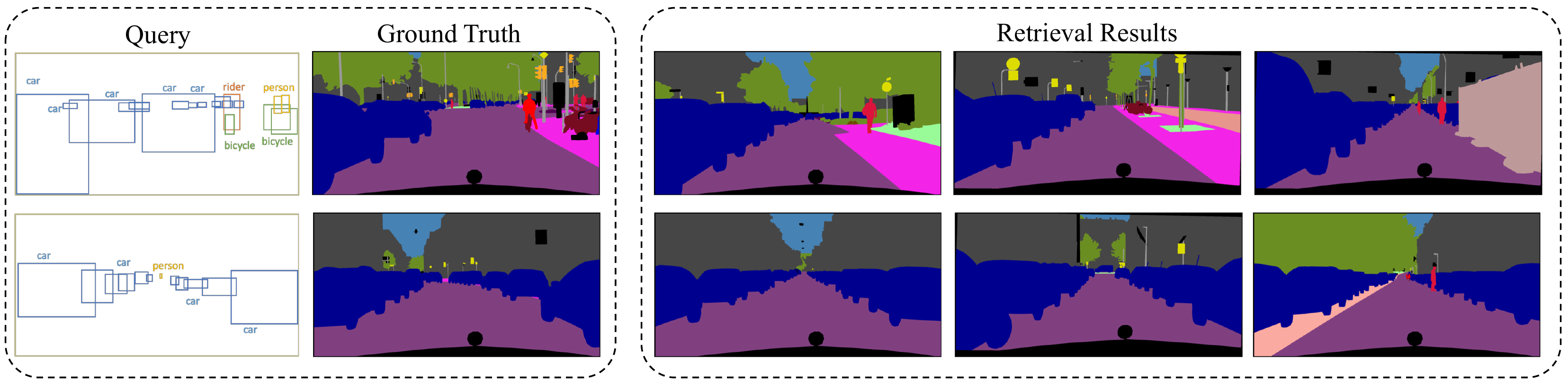}
  \vspace{-2mm}
  \caption{Examples of top-3 retrieved images from the memory bank of Cityscapes.}
  \label{fig:vis:retrive}
\end{figure*}

\vspace{5pt}
\noindent \textbf{Implementation Details} \, All the experiments are conducted on an NVIDIA DGX1 with 8 V100 GPUs. We use Adam \cite{adam} as the optimizer, and learning rates for the generator and discriminator are both set to $0.0002$. For Cityscapes, we train 60 epoches to obtain a good generator, and ADE20k needs 150 epoches to converge. $m$ is set to 3 for both datasets.

\subsection{Quantitative Evaluation}
Table \ref{tab:fid} summarizes the results of all the models w.r.t. the FID score and classification accuracy. We also report the scores on images generated from vanilla SPADE (SPADE-v) using segmentation map as input (upper bound). Measured by FID, BachGAN outperforms all the baselines in both datasets with a relatively large margin. For Cityscapes, BackGAN achieves a FID score of 73.3, which is close to the upper bound. In ADE20K, the improved gain over baselines is not significant. This is because most images in ADE20K are dominated by salient foregrounds, with relatively less space for background, which limits the effect of our hallucination module. The pixel accuracy of our method is also higher than other baselines. BachGAN-r also achieves reasonable performance on both datasets.

\subsection{Qualitative Analysis}
In Figure \ref{fig:vis:city} and \ref{fig:vis:ade20}, we provide qualitative comparison of all the methods. Our model produces images with much higher visual quality compared to the baselines. Particularly, in Cityscapes, our method can generate images with detailed/sharp backgrounds while the other approaches fail to. In ADE20K, though the background region is relatively smaller than Cityscapes, BachGAN still produces synthesized images with better visual quality. 

Figure \ref{fig:seq_edit_cityscape} and \ref{fig:seq_edit_ade20k} demonstrate that BachGAN is able to manipulate a series of complex images progressively, by starting from a simple layout and adding new bounding boxes sequentially.
The generated samples are visually appealing, with new objects depicted at the desired locations in the images, and existing objects remain consistent to the layout in previous rounds. These examples demonstrate our model's ability to perform controllable image synthesis based on layout. 

Figure~\ref{fig:obj_flip} further illustrates that BachGAN also works well when objects are positioned in an unconventional way. In the bottom row, we flip some objects in the object layouts (\emph{e.g.}, windowpane in the top-left image), and generate images with the manipulated layouts. BachGAN is still able to generate high-quality images with a reasonable background, proving the robustness of BachGAN.

In Figure \ref{fig:vis:retrive}, we sample some retrieved images from Cityscapes. The top-3 results are consistent and similar with the original background. More synthesized and retrieval results (for ADE20K) are provided in Appendix.

\begin{table*}[t!]
\begin{center}
\small
\begin{tabular}{c|ccc|ccc|ccc|ccc}
\hline
\multirow{2}{*}{Dataset} & \multicolumn{3}{c|}{BachGAN vs.} & \multicolumn{3}{c|}{BachGAN vs.} & \multicolumn{3}{c|}{BachGAN vs. } & \multicolumn{3}{c}{BachGAN vs.} \\
& \multicolumn{3}{c|}{SPADE} & \multicolumn{3}{c|}{SPADE-Seg} & \multicolumn{3}{c|}{BachGAN-r} & \multicolumn{3}{c}{Layout2im} \\
\hline \hline
& win & loss & tie & win & loss & tie & win & loss & tie & win & loss & tie\\ \hline
Cityscapes & 85.5 & 3.4 & 11.1 & 71.7 & 12.4 & 15.9 & 61.6 & 24.1 & 14.3 & 96.0 & 0.2 & 3.8 \\
\hline
ADE20K & 75.9 & 12.8 & 11.3 & 66.8 & 17.4 & 15.8 & 57.2 & 18.7 & 24.1 & - & - & -\\
\hline
\end{tabular}
\end{center}
\vspace{-5mm}
\caption{User preference study. Win/lose/tie indicates the percentage of images generated by BachGAN are better/worse/equal to the compared model.}
\label{tab:human_eval}
\end{table*}

\subsection{Human Evaluation} 

We use Amazon Mechanical Turk (AMT) to evaluate the generation quality of all the approaches. AMT turkers are provided with one input layout and two synthesized outputs from different methods, and are asked to choose the image that looks more realistic and more consistent with the input layout. 
The user interface of the evaluation tool also provides a neutral option, which can be selected if the turker thinks both outputs are equally good. We randomly sampled 300 image pairs, each pair judged by a different group of three people. Only workers with a task approval rate greater than 98\% can participate in the study.

Table \ref{tab:human_eval} reports the pairwise comparison between our method and the other four baselines. Based on human judgment, the quality of images generated by BachGAN is significantly higher than SPADE. Comparing with two strong baselines (SPADE-SEG and BachGAN-r), BachGAN achieves the best performance. As expected, Layout2im receives the lowest acceptance by human judges, due to its low resolution.

\begin{table}[t]
\centering
\begin{tabular}{c|c|c|c}
\hline 
Method & BachGAN-3 & BachGAN-4 & BachGAN-5 \\
\hline
FID & 73.31 & 73.03 & 72.95\\
\hline
\end{tabular}
\vspace{-3mm}
\caption{FID scores of BachGAN with different numbers of retrieved segmentation maps (Cityscapes).}
\label{tab:size:memory}
\end{table}

\begin{table}[t!]
\centering
\begin{tabular}{c|c|c}
\hline 
Bank size  & BachGAN & BachGAN-r \\
\hline
$|\mathbf{B}|$ & 73.31 & 74.44\\
\hline
$2 \times |\mathbf{B}|$ & 72.50 & 73.95\\
\hline
\end{tabular}
\vspace{-3mm}
\caption{FID scores of BachGAN and BachGAN-r trained using memory bank of different sizes (Cityscapes).}
\label{tab:size:bank}
\end{table}

\subsection{Ablation Study}

\noindent \textbf{Effect of segmentation map retrieval} \, First, we train three BachGANs with different numbers of retrieved segmentation maps, setting $m$ to 3, 4 and 5, and evaluate them on Cityscapes. The FID scores of different models are summarized in Table \ref{tab:size:memory}. The model using Top-5 retrieved segmentation maps (BachGAN-5) achieves the best performance, compared to models with Top-3 and Top-4. This analysis demonstrates that increasing the number of selected segmentation maps can slightly improve the scores. Due to the small performance gain, we keep $m=3$ in our experiments.

\vspace{5pt}
\noindent \textbf{Effect of memory bank} \, We also compare models trained with memory banks of different sizes. Specifically, we compare the performance of BachGAN and BachGAN-r with memory bank size $|\mathbf{B}|$ (used in our experiments) and $2 \times |\mathbf{B}|$. Results are summarized in Table \ref{tab:size:bank}. With a larger memory bank, both models are able to improve the evaluation scores. Interestingly, the gain of BachGAN is larger than that of BachGAN-r, showing that BachGAN enjoys more benefit from the memory bank. More analysis about the size of memory bank is provided in Appendix. 

\section{Conclusion}
In this paper, we introduce a novel framework, BachGAN, to generate high-quality images conditioned on salient object layout. By hallucinating the background based on given object layout, the proposed model can generate high-resolution images with photo-realistic foreground and integral background.  Comprehensive experiments on both Cityscapes and ADE20K datasets demonstrate the effectiveness of our proposed model, which can also perform controllable image synthesis by progressively adding salient objects in the layout. 
For future work, we will investigate the generation of more complicated objects such as people, cars and animals \cite{psgan}. Disentangling the learned representations for foreground and background is another direction.

\vspace{5pt}
\noindent \textbf{Acknowledgements} \,
This work was also supported in part by NSF-1704309.

\clearpage 
\appendix
\newpage
\begin{figure*}[t!]
  \centering
  \includegraphics[width = \textwidth]{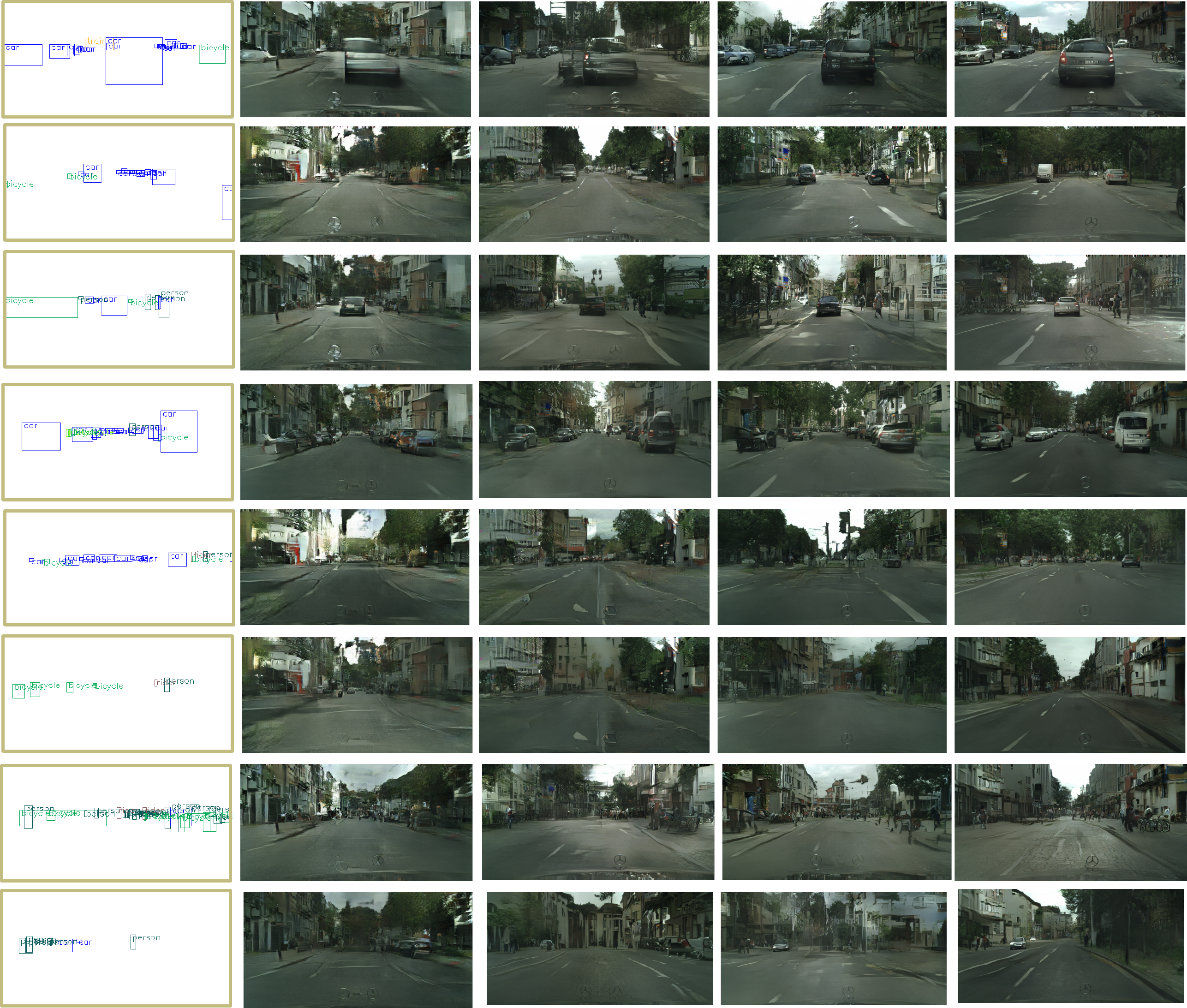}
  \caption{Additional examples of image synthesis results from different models on the Cityscapes dataset. }
  \label{fig:vis:city:app}
\end{figure*}

\begin{figure*}[t]
  \centering
  \includegraphics[width = 0.98\textwidth]{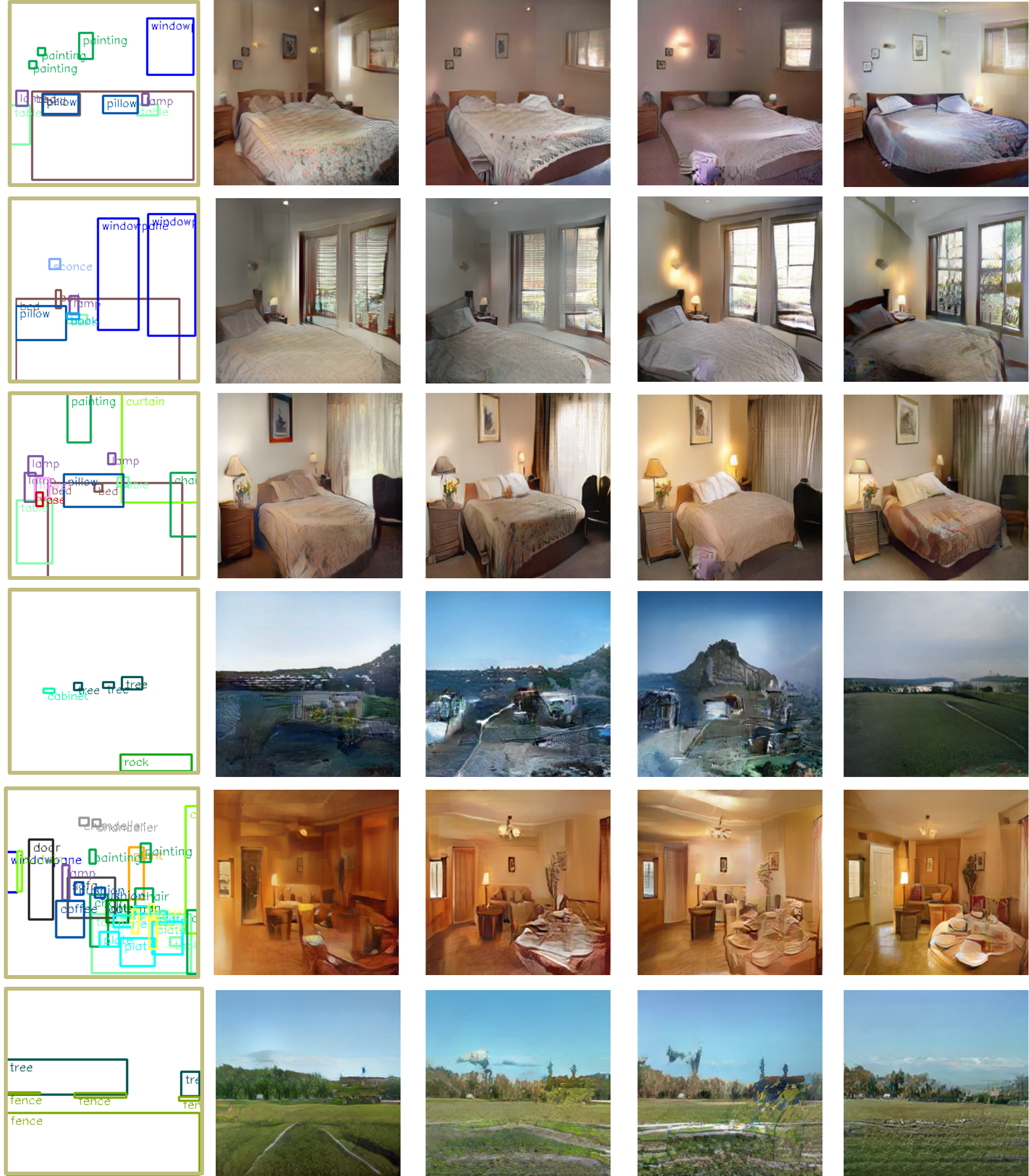}
  \caption{Additional examples of image synthesis results from different models on the ADE20K dataset. }
  \label{fig:vis:ade20:a}
\end{figure*}

\begin{figure*}[t]
  \centering
  \includegraphics[width = 0.98\textwidth]{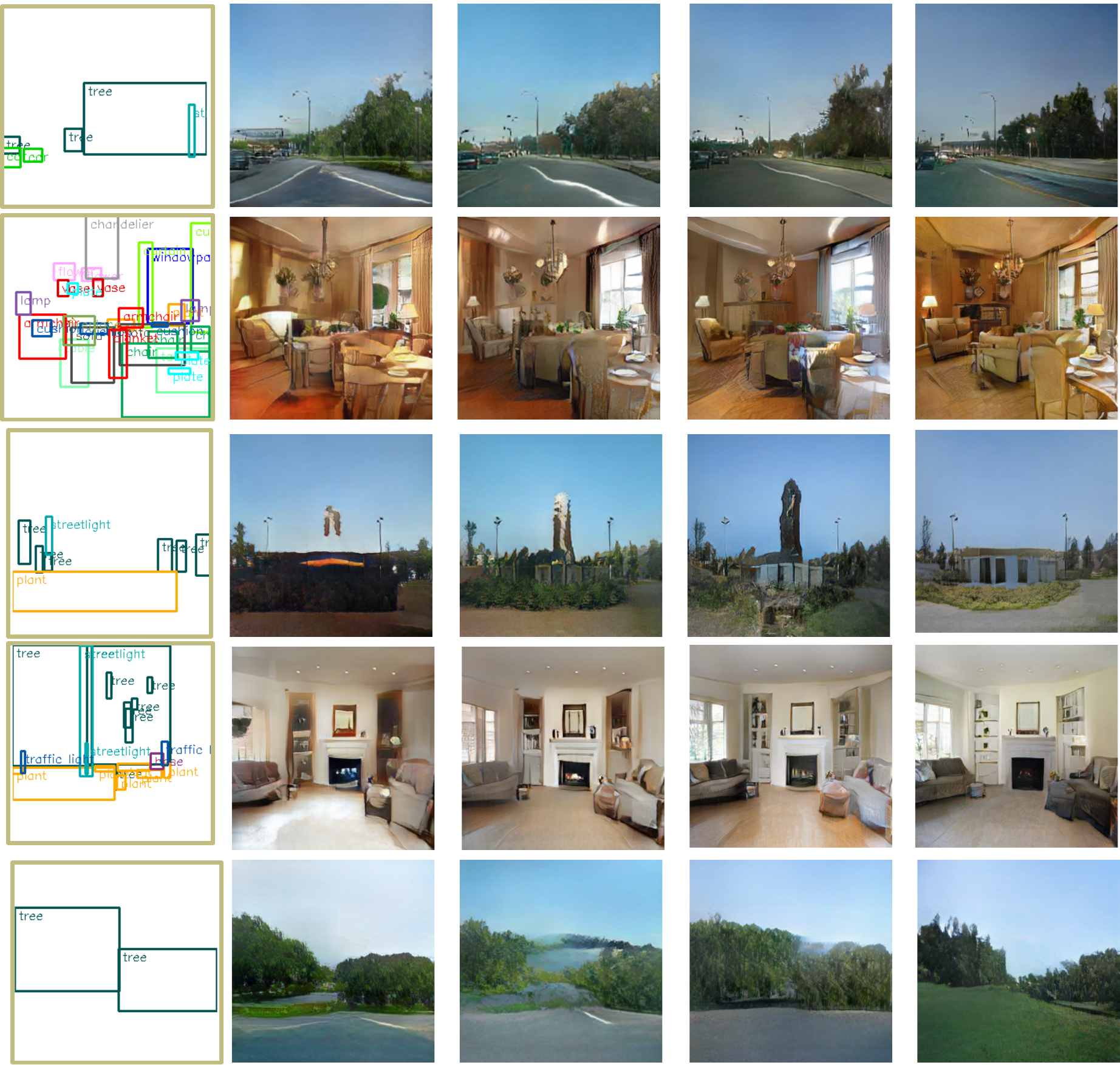}
  \caption{Additional examples of image synthesis results from different models on the ADE20K dataset.}
  \label{fig:vis:ade20:b}
\end{figure*}

\begin{figure*}
  \centering
  \includegraphics[width = 0.98\textwidth]{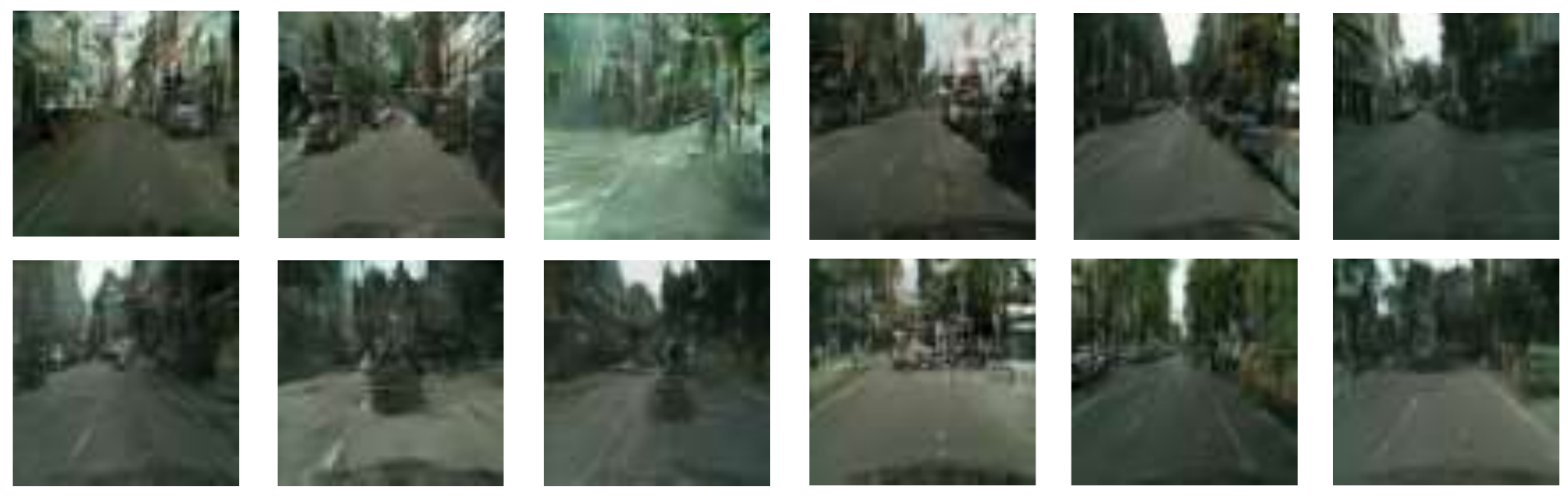}
  \caption{Examples of image synthesis results from Layout2im on the Cityscapes datasets.}
  \label{fig:vis:layout2im}
\end{figure*}

\section{Additional Ablation Study}
In Table \ref{tab:size:bank:app}, we show additional results of models trained with memory banks of different size on both datasets. With half-sized memory banks, BachGAN still achieves good performance measured by FID scores. These results demonstrate that our model is not sensitive to this hyper-parameter. 

\begin{table}[h]
\centering
\begin{tabular}{c|c|c}
\hline 
Datasets  & Cityscapes & ADE20K \\
\hline
$\frac{1}{2} \times |\mathbf{B}|$ & 74.2 & 51.2 \\
\hline
$|\mathbf{B}|$ & 73.3 & 49.8\\
\hline
$2 \times |\mathbf{B}|$ & 72.5 & -\\
\hline
\end{tabular}
\caption{FID scores of BachGAN trained using different memory bank sizes on Cityscapes and ADE20K.}
\label{tab:size:bank:app}
\end{table}

\section{Additional Implementation Details}
\paragraph{Generator} The architecture of the generator follows the one used in~\cite{park2019semantic}, which consists of several SPADE Residual Blocks with nearest neighbor upsampling. We train the model on 8 V100 GPUs with synchronized BatchNorm. Instead of feeding the label map of object layout into each SPADE layer, we obtain the object layout with fine-grained background (or the fusion of several background segments), and take it as the input. 

\paragraph{Discriminator}
The architecture of the discriminator also follows the one used in SPADE ~\cite{park2019semantic} and pix2pixHD~\cite{wang2018high}, which
uses a multi-scale design with InstanceNorm (IN), and applies Spectral Norm to all the convolutional layers. Compared with them, the difference of our model is that the input includes both real images and the object layouts with hallucinated background.

\paragraph{Background Fusion Module}
As illustrated in Figure 2 and Section 3.3 of our main manuscript, the Background Fusion Module has 2 convolutional neural networks -- $\mathcal{F}$ and  $\mathcal{M}$.  Specifically, $\mathcal{F}$ is used for encoding the label maps, and $\mathcal{M}$ is used to process the fused feature of foreground layout and background segments. Both $\mathcal{F}$ and $\mathcal{M}$
include a $3 \times 3$ convolutional layer with $k$ convolutional filters and ReLU activation functions, where $k$ is $C_o+C_b$. 

\paragraph{Retrieval Time} The average time for calculating each $\text{IoU}_r$ is 4ms. In our paper, we compute all $\text{IoU}_r$ in a sequential way (one by one), which takes $\sim$20 seconds (for $|\mathbf{B}|$ = 5,000 maps).
This could be further improved by paralleling the computation.

\paragraph{Sequential Generation} In order to generate consistent images, we use the same retrieved semantic segmentation maps as the background inputs. In addition, the generation process is deterministic, which means no random noise is added to the training and testing stages.This property indicates that BachGAN can generate similar results for each object as long as they have the same layout.

\section{Additional Synthesized Images}
In Figure \ref{fig:vis:city:app}, \ref{fig:vis:ade20:a} and \ref{fig:vis:ade20:b}, we show additional synthesized images from the proposed method on the Cityscapes and ADK20K datasets, with comparisons to the baseline models. 
In Figure \ref{fig:vis:layout2im}, we show the generated images using Layout2im \cite{zhaobo2019layout2im} on Cityscapes. 
\end{document}